\documentclass[9pt,twocolumn,twoside]{osajnl}

\journal{jocn} 

\setboolean{shortarticle}{false}

\title{Deep learning for enhanced free-space optical communications}

\author[1,*]{Manon P. Bart}
\author[1]{Nicholas J. Savino}
\author[2]{Paras Regmi}
\author[3]{Lior Cohen}
\author[4]{Haleh Safavi}
\author[4]{Harry C. Shaw}
\author[5]{Sanjaya Lohani}
\author[5]{Thomas A. Searles}
\author[1,6]{Brian T. Kirby}
\author[2]{ Hwang Lee}
\author[1,**]{Ryan T. Glasser}

\affil[1]{Department of Physics and Engineering Physics, Tulane University, New Orleans, Louisiana 70118, USA}

\affil[2]{Department of Physics and Astronomy, Louisiana State University, Baton Rouge, Louisiana 70803, USA}

\affil[3]{Department of Electrical, Computer and Energy Engineering, University of Colorado Boulder, Boulder, Colorado 80309, USA}

\affil[4]{NASA Goddard Space Flight Center, Greenbelt, Maryland 20771, USA}

\affil[5]{Department of Electrical \& Computer Engineering, University of Illinois Chicago, Chicago, Illinois 60607, USA}

\affil[6]{DEVCOM Army Research Laboratory, Adelphi, MD 20783, USA}

\affil[*]{Corresponding author: mbart1@tulane.edu}

\affil[**]{Corresponding author: rglasser@tulane.edu}

\begin{abstract}
Atmospheric effects, such as turbulence and background thermal noise, inhibit the propagation of coherent light used in ON-OFF keying free-space optical communication. Here we present and experimentally validate a convolutional neural network to reduce the bit error rate of free-space optical communication in post-processing that is significantly simpler and cheaper than existing solutions based on advanced optics. Our approach consists of two neural networks, the first determining the presence of coherent bit sequences in thermal noise and turbulence and the second demodulating the coherent bit sequences. All data used for training and testing our network is obtained experimentally by generating ON-OFF keying bit streams of coherent light, combining these with thermal light, and passing the resultant light through a turbulent water tank which we have verified mimics turbulence in the air to a high degree of accuracy. Our convolutional neural network improves detection accuracy over threshold classification schemes and has the capability to be integrated with current demodulation and error correction schemes.
\end{abstract}

\setboolean{displaycopyright}{true}

\begin{document}

\maketitle

\section{Introduction}
Optical communication has become increasingly widespread due to its ability to support a larger bandwidth, narrower beam divergence, and smaller weight and power requirements than radio frequency links \cite{FSO}. With these technical advantages,  free-space optical (FSO) communication has become a prominent approach for meeting the technological requirements of long-distance and space-based communications \cite{SON201767}. In addition, the availability of state-of-the-art commercial optical devices and recent demonstrations in ground-to-satellite, satellite-to-satellite, and satellite-to-ground communication schemes have confirmed the advantages of optical communication systems \cite{ARTO,NASA,NASA1}. 

Free-space optical communication is currently implemented through various modulation schemes, one of the most prominent being ON-OFF keying (OOK), due to its simplicity \cite{ook1,ook2}. OOK is a commonly used form of amplitude shift keying modulation, where light pulses are sent in binary bits that are either ``ON'' (1) or ``OFF'' (0). However, as is the case with other large scale continuous optical communication schemes, OOK schemes are susceptible to fluctuations from atmospheric transmission \cite{LSUthermalpaper}. Optical loss increases exponentially over long-distances which leads to degradation of transmitted pulses \cite{Olexp}. Additional loss can also occur due to turbulence in the atmosphere, cloud coverage, and varying weather conditions which all play a role in pulses arriving at the detector being attenuated and noisy. \cite{Bphot,FSO,LSUthermalpaper,AT}. These losses lead to errors in the demodulation of coherent bit streams, which ultimately degrades the bit-error-rate (BER). Recently, deep learning has become a popular approach to mitigating the effects of turbulence and other sources of error in FSO systems. Convolutional neural networks (CNNs) are a class of artificial neural networks which are often used to analyze and characterize visual imagery. Given a noisy input, the neural network can analyze repetitive features from large datasets to classify these inputs. Previous work has shown the efficacy of neural networks and deep learning in improving free-space optical communication schemes \cite{App1, App2, App3,App4,Lohani2,Lohani1,app7,app8,app9,lohani2020coherent}. In recent years, CNNs have been used in particular to increase the information capacity of optical communication schemes using Laguerre-Gaussian modes and various demodulation techniques \cite{app5,app6,app10,app11}. 

Here we develop and experimentally validate a custom CNN framework for classifying sequences of OOK bits at various signal-to-noise ratios (SNRs) in the presence of thermal noise and turbulence. We experimentally produce thermal noise and turbulence-like effects to model the process of sending OOK bits through the atmosphere, which we use to train and test our model. First, we classify whether or not portions of detected thermal light contain bit streams. Then, once we detect the presence of bits, we classify individual bits as ``ON'' (1) or ``OFF'' (0). The work aims to use CNNs to create high-throughput optical communication schemes in noisy environments. The results presented here demonstrate the utility of neural networks as valuable tools in FSO systems with or without additional optics-based error correction schemes. 

\section{Experimental Methods}
\subsection{Experimental Setup}

\begin{figure}[htbp]
    \centering
    \includegraphics[width=\linewidth]{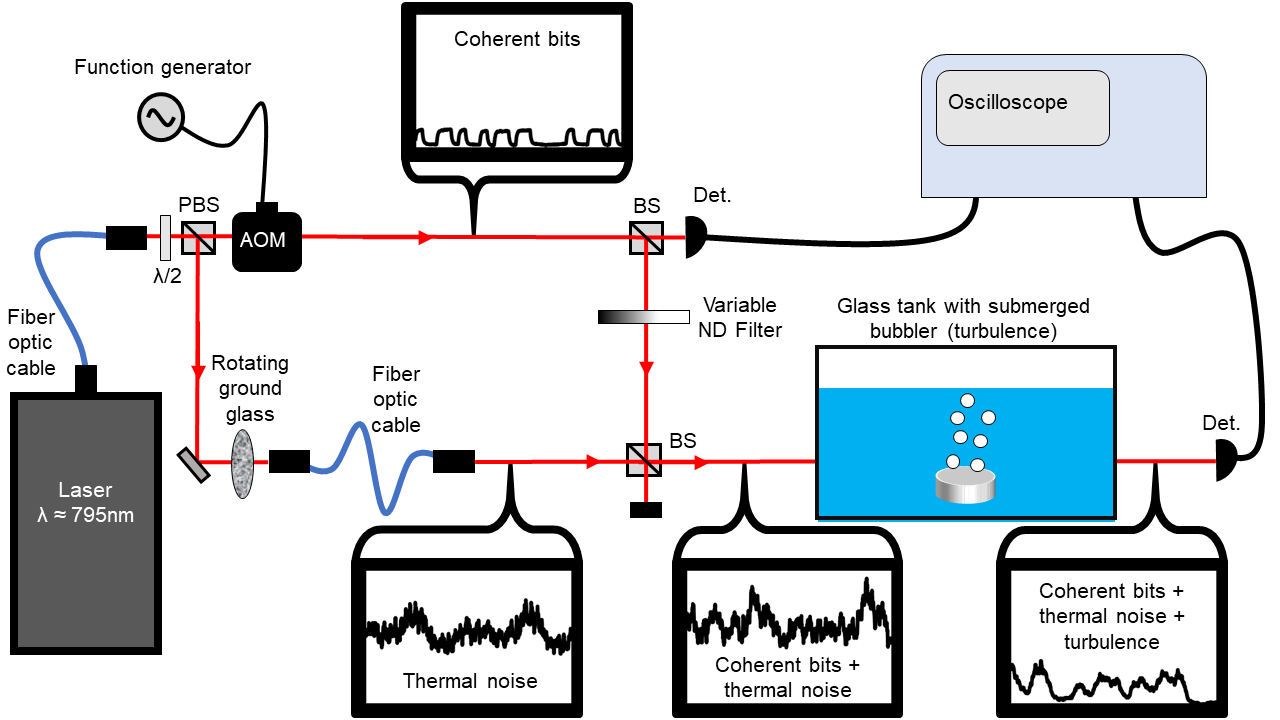}
    \caption{Diagram of the experimental setup, with plots of light intensity at several points. Light comes from a coherent amplified diode laser source ($\approx$ 795 nm). The beam is split such that some coherent light passes through an acousto-optic modulator, generating a pattern of coherent bits, while the other part of the beam passes through rotating ground glass and a fiber optic cable, creating pseudo-thermal light. The two seperate beams are recombined on a beamsplitter, then passed through turbulence. Abbreviations: BS = beamsplitter, AOM = acousto-optic modulator, Det. = photodetector, PBS = polarized beamsplitter, $\lambda/2$ = half waveplate, ND = neutral density, BS = 50:50 non-polarizing beamsplitter.}
    \label{setup}
\end{figure}

A near-infrared ($\lambda \approx 795$nm) laser is used to separately generate  coherent bits and thermal noise. The output of the laser is split by a polarizing beamplitter such that the transmitted beam becomes the coherent bit stream and the reflected beam becomes the thermal noise. We generate pseudo-thermal light by passing the coherent light through rotating ground glass. The light is focused onto the ground glass with a f = 100 mm lens, and then re-collimated with a f = 25.5 mm lens, and coupled into a fiber optic cable to sample the generated speckle pattern. The OOK bit stream is generated by passing the transmitted beam through an acousto-optic modulator controlled by a function generator, resulting in a pattern of bits, each 3.5 ms in duration.  The coherent bit stream is then split with a 50:50 non-polarizing beamsplitter, with half of the light going directly to a photodetector in order to determine the actual OOK bit stream, while the other half of the coherent bits are combined with the pseudo-thermal light on a 50:50 non-polarizing beamsplitter. We then pass the combination of light through turbulent water in a tank. To create turbulence in the water, the beam is passed through bubbles generated by a submerged airstone, often referred to as a `bubbler.' The light intensity is then detected with a photodetector. A schematic diagram is shown in Figure \ref{setup}.

We measure the intensity of the detected light with the photodetectors and an oscilloscope. The oscilloscope takes data points every .1 ms; each ``ON'' or ``OFF'' bit is 35 values long. In addition, beam blocks are placed at various point to generate different conditions for the light, or to turn the bubblers off to examine the light without turbulence. We vary the amplitude (and thus SNR) of the coherent light by adjusting the variable neutral density filter.

\subsection{Channel Emulation}

In order to emulate a channel analogous to atmospheric transmission, we study the statistics of the turbulence created by the bubbler as well as provide a large range of signal-to-noise ratios congruous to a range of weather conditions and transmitter intensities. 
Turbulence can be caused by variation of temperature and pressure of the atmosphere along the propagation path which causes constructive or destructive interference of the propagating beam. These random fluctuations in the intensity of the received signal is known as scintillation. The statistics of  turbulence have been heavily studied \cite{SC1,SC2,SC3}. Turbulence is often modeled with log-normal (weaker turbulence), negative exponent (stronger turbulence), or gamma-gamma distributions (moderate to strong turbulence) \cite{FSO}. Log-normal turbulence is modeled by the distribution $f(I)$, over the intensity $I$, where

\begin{equation}
    f(I) = \frac{1}{2I\sqrt{2\pi\sigma^{2}_{I}}}\exp\left(-\frac{\left(\ln\frac{I}{I_{0}}+\frac{\sigma^{2}_{I}}{2}\right)^{2}}{2\sigma^{2}_{I}}\right).
\end{equation}
Here, $I_{0}$ is the mean irradiance and $\sigma^{2}_{I}$ is the scintillation index \cite{7968506,trichili2021retrofitting}.

\begin{figure}[ht!]
    \centering
    \includegraphics[scale=0.18]{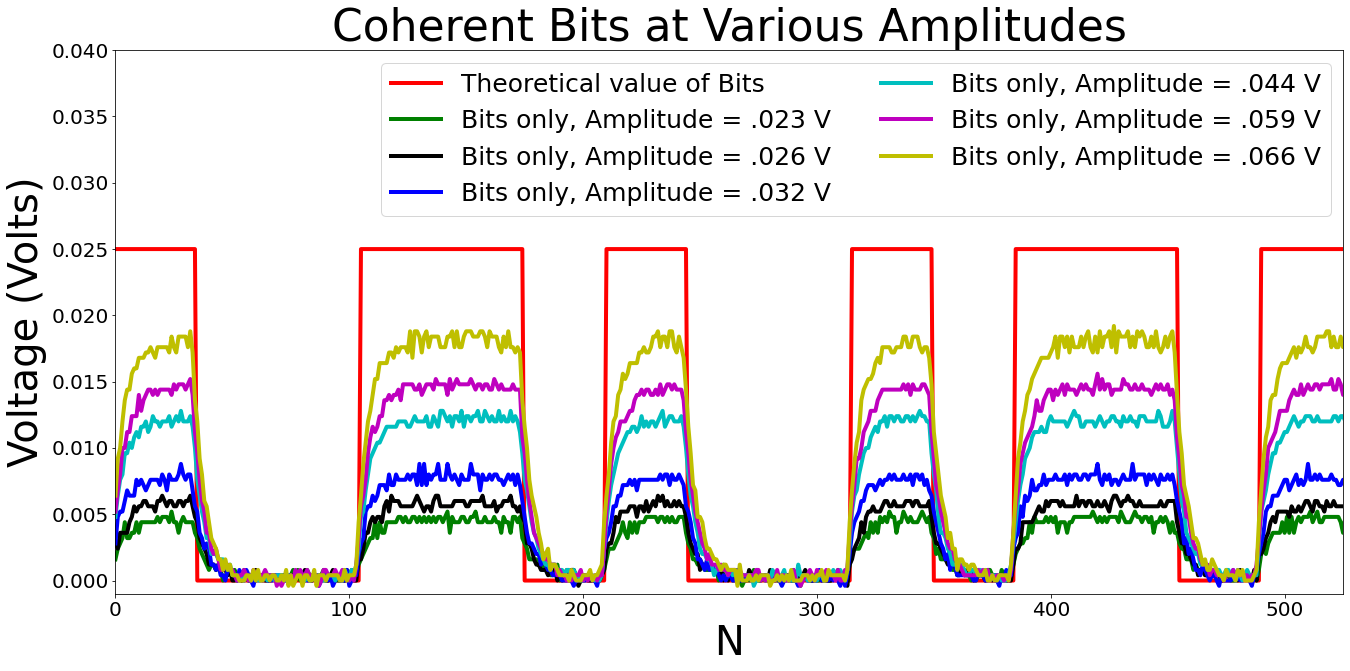}
    \caption{Coherent OOK bit sequences that are not in the presence of thermal noise and turbulence for various signal-to-noise ratios. One coherent bit pulse is defined by an increment of 35 values. N represents these consecutive values we receive, from N = 0 to 525, which is equivalent a bit sequence of length 15. The red solid line depicts the theoretical value of the bits being ``ON'' (1)  or ``OFF'' (0). The y axis indicates the detected voltage on a photodiode. Analysis methods examined later in this paper are not dependant on the units or a global scale factor, so other types of detectors and other units may be used without issue.}
    \label{CohOut}
\end{figure}

The noise generated by passing light through the bubbles in the water tank creates fluctuations in intensity and follows studied turbulence distributions. The statistics of underwater turbulence has been studied, and in addition, relationships between atomspheric and underwater turbulence have been established \cite{turb1,turb2,turb3}. The experimental scintillation index is in the range of accepted values for that of atmospheric turbulence \cite{turbdayton,wayne_2010}. Via a curve fit, we observe a scintillation index of $1.8 \pm .2$ throughout the experiment. The efficacy of our neural network is also studied for a wide range of SNRs. The ranges of coherent bits passed through the experimental set up are shown in Figure \ref{CohOut}. N is the number of consecutive values received by the detector, where 35 consecutive values is equivalent to one bit. The lowest SNR is comparable to a signal in a high noise environment, such as bits in the presence of heavy cloud coverage.

\subsection{Neural Network}
One of the successes of neural networks is the focus on pattern recognition in data rather than statistical signal processing, where the neural network uses deep learning algorithms to discover latent information which will aid in classification. In the case of classifying bit sequences, large sets of data where the bit pattern is known are fed into a neural network which is able to recognize the patterns associated with the presence of coherent bits in order to demodulate incoming signals. This is particularly useful for signal demodulation in high noise environments where the statistics of the environment are complex \cite{Lohani2}. Following the experimental data generation, the data is preprocessed in order to be input into the neural network. 

\subsubsection{Pre-Processing}

\begin{figure}[ht!]
    \centering
    \includegraphics[width=.95\linewidth]{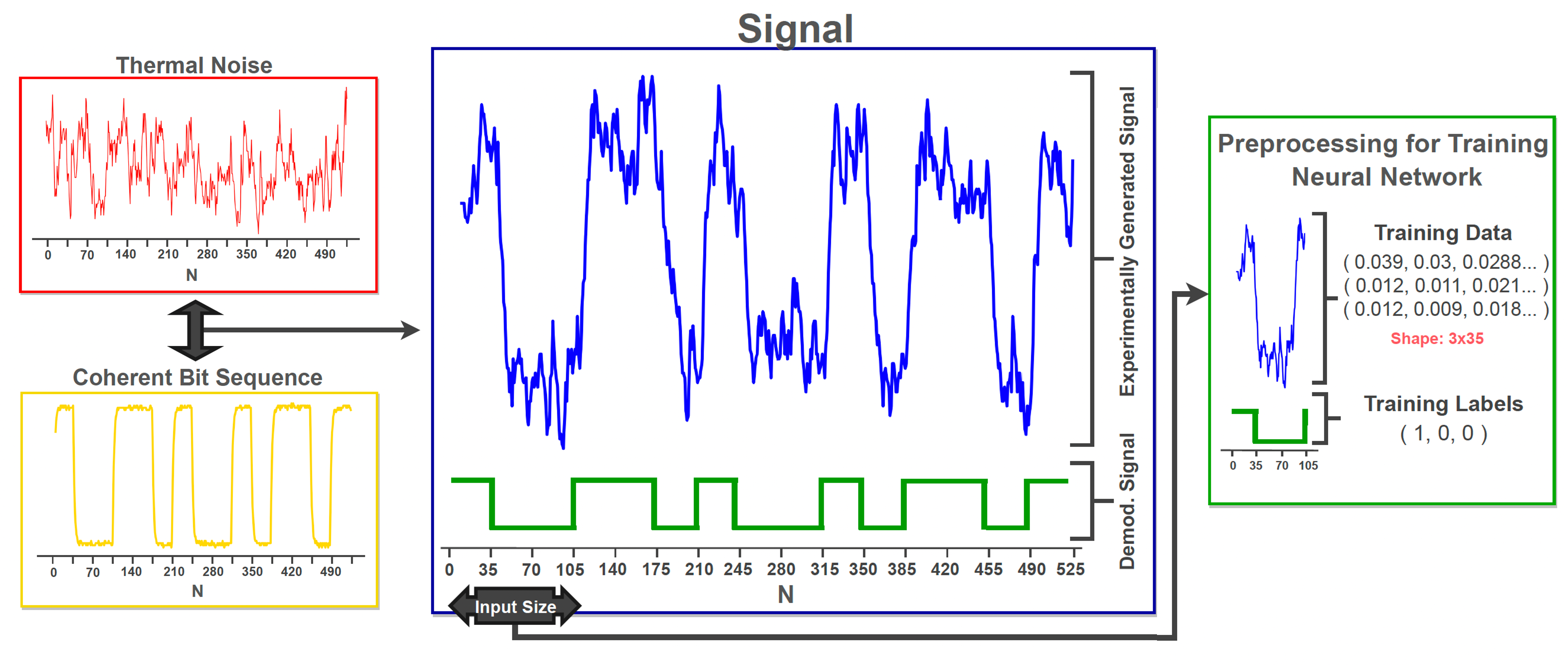}
    \caption{Pre-processing of the neural network. We begin with coherent bits and thermal noise/thermal noise and turbulence. N is defined as the consecutive values generated, where 35 consecutive values corresponds to 1 bit. The bit sequence and noise is combined to create the experimentally generated signal in blue. The green corresponds to the known demodulated signal. The experimentally generated signal is spliced into various input sizes which are converted into matrices. These are used as training data. The green signal is used as training labels which correspond to whether the bit is "ON" or "OFF". In this image, the three bits are input into the neural network, where their amplitudes are reshaped into a 35 by 3 matrix. For training, the 3 bits are labeled appropriately as "ON", "OFF", "OFF". By utilizing supervised learning, the neural network will begin recognizing consecutive amplitude values that correspond with bits being "ON" and "OFF".  The units here are intensity in arbitrary units, as the neural network focuses on the on the relative amplitude values when classifying.} 
    \label{Pre}
\end{figure}

In order to train our CNN, we must first perform pre-processing on experimental data to create a labeled set of examples on which to perform supervised learning.
The data consists of individual bits which contain N = 35 consecutive values, which corresponds to the amplitude of the coherent bit and noisy background. Additionally, the bits are labeled as to whether or not coherent bits are present. The consecutive bit amplitudes and whether or not they are ``ON'' or ``OFF'' corresponds to the training data and training labels, respectively. An outline of our approach for generating training data is shown in Figure \ref{Pre}.

The experimentally generated signal, which includes noise and coherent bit sequences and the previously known demodulated signal, is input into the neural network. The number of consecutive values per input varies, where the number of consecutive values is divisible by 35 to capture complete bits. The bit value intensity is then reshaped into a 35 by "x" matrix, where "x" corresponds to the number of bits that will be put into the neural network at once. Finally, the known demodulated signal is used to label the training set. 

After training, unlabeled data acts as a validation set to check the neural network's ability to classify the data. The experimentally generated data is split into 80\% for training and 20\% for validation sets, which leads to around 2,000,000 bits used to train the data set. The results presented in the paper correspond to validation set accuracy, and as the CNN is pre-trained, the validation set accuracy is analogous to classification accuracy for real-time implementation. For the first neural network, we classify whether or not there are any coherent bits in a sequence. Following the first neural network, sequences where coherent bits are detected are input into the second neural network which classifies each individual bit as "ON" or "OFF". 

\subsubsection{Network Architecture}

\begin{figure}[ht!]
    \centering
    \includegraphics[width=\linewidth]{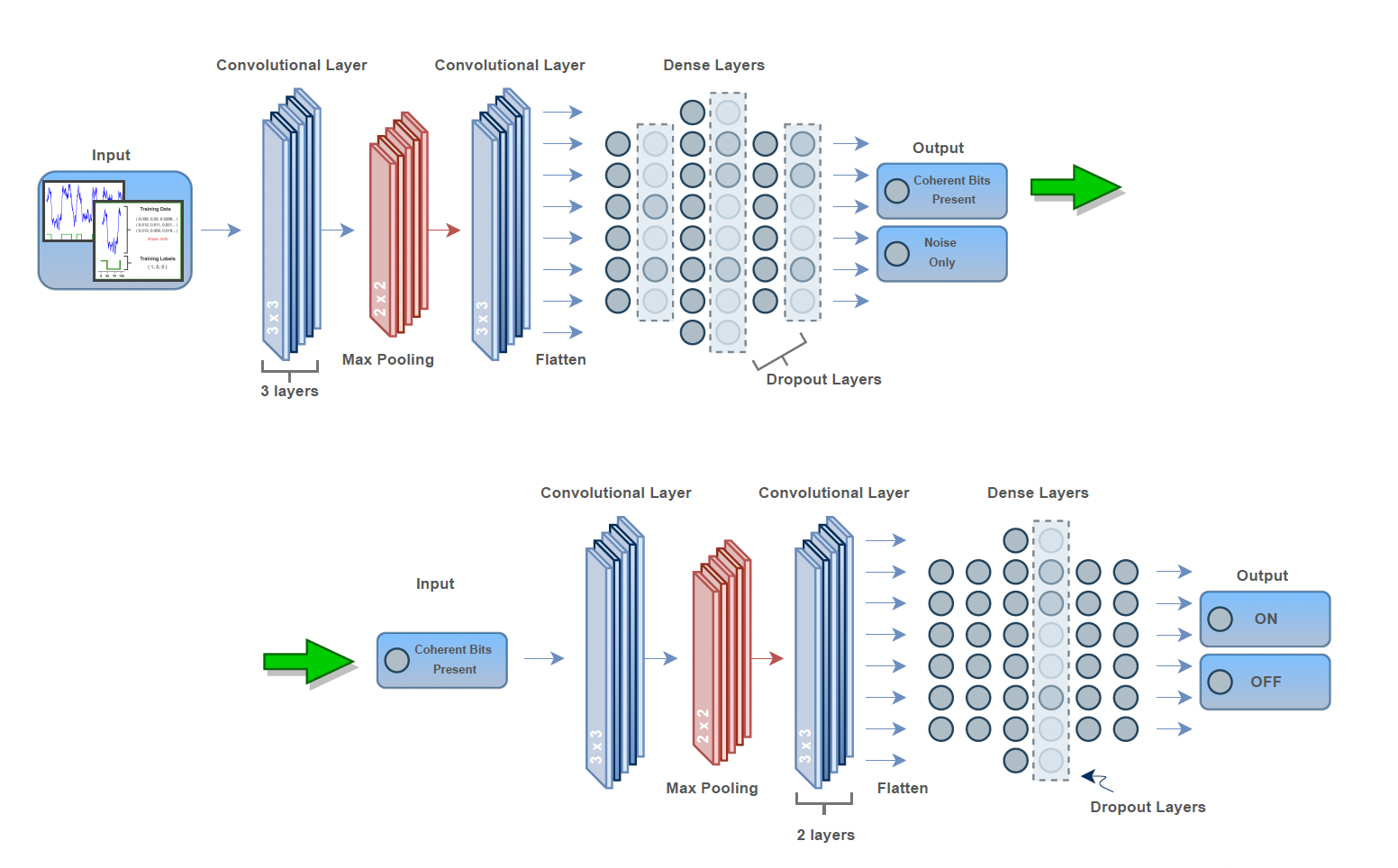}
    \caption{The convolutional neural network (CNN) architecture. Noisy data is input into the CNN and outputs whether coherent OOK bits are present or just thermal noise/thermal noise and turbulence. The data was split into 80/20 \% for the training and validation set, respectively. Data is first fed into three convolutional layers followed by a max pooling layer and another convolutional layer. The data is then fed into dense layers and dropout layers which aid in overfitting. The final dense layer is the binary output which indicates presence or absence of an OOK bit stream in a given sequence. Following the first neural network, the coherent OOK bit streams detected are input into the convolutional neural network and output whether or not the bits are ``ON'' (0) of ``OFF'' (1). The CNN is similar in layout to the first. }
    \label{NN1}
\end{figure}

Our network architecture consists of a two-stage approach where we first classify input data as "solely background noise" or "includes OOK bits," with the latter resulting in an additional stage where a second network attempts to demodulate the bit stream. The first neural network attempts to detect the presence of a bit stream before demodulation in order to mitigate computational time if no bit streams are present. The second neural network is trained to recognize the signal's features at the receiver that correlate to a coherent OOK bit pulse, allowing our model to classify bits as ``ON'' or ``OFF.'' The scope of our work varies from high SNRs to very low SNRs, where accurate classification is impossible without using advanced demodulation and error correction techniques. A schematic of the CNN based demodulation scheme is shown in Figure \ref{NN1}. 

The first neural network consists of three convolutional layers, followed by a maxpooling layer and another convolutional layer. Following the convolutional neural network, the data is input into dense layers with dropout layers in between. Afterwards, the data set leads to a final dense layer with a binary output of the presence of coherent OOK bit pulses in noise or just thermal noise/thermal noise and turbulence. Several different algorithms have been suggested for adaptive optic schemes including stochastic-gradient-descent and Gerchberg-Saxton \cite{alg1,alg2,alg3}. However, we find optimal classification accuracy and reduced computational time using the Adam optimizer, which is an extension of stochastic gradient descent. The Adam optimizer is used with a learning rate of 0.001 and the cross entropy loss function is used. This neural network is run for 50 epochs to reach a high classification accuracy. 

After identifying the OOK bit stream, a second neural network classifies the bit pulses as ``ON'' or ``OFF,'' demodulating the signal. The neural network has a similar configuration to the first neural network. The data is reshaped into single bit values of length 35 which results in approximately 500,000 bits in total. These bits are passed through the convolutional and maxpooling layers and then into dense layers with dropout layers present as outlined in Figure \ref{NN1}. The dropout layer had a rate of 60 $\%$ and the layers used a rectified linear unit (ReLU) activation function. We optimized the hyperparameters and our resulting CNN ran for 20 epochs. The Adam optimizer is used again, as well as the Categorical Cross entropy loss function. For both the first and second neural network, we divide the training of the neural network and the testing in an 80\% and 20\% split, respectively. By using a high percentage of data for training, we can create a finely-tuned classification tool which will be trained for use in real time. In order to combat overfitting, we incorporate dropout layers in both neural networks and tuned the architecture to our final result which insured high accuracy for the testing set.

\section {Results}
Currently, many tools aid in signal processing, and some of the more widely used demodulation techniques include adaptive optics, background noise rejection, relay transmission, and hybrid RF/FSO communications \cite{FSO}. While techniques such as adaptive optics aid in demodulating noisy signals, neural networks have several advantages that we explore in this paper. One of the key advantages of using a neural network is that it is integrable into other tools used to demodulate signals, such as adaptive optics, while not significantly increasing the system's size, weight, and power. 

\begin{figure*}[ht!]
    \centering
    \includegraphics[width=.8\textwidth]{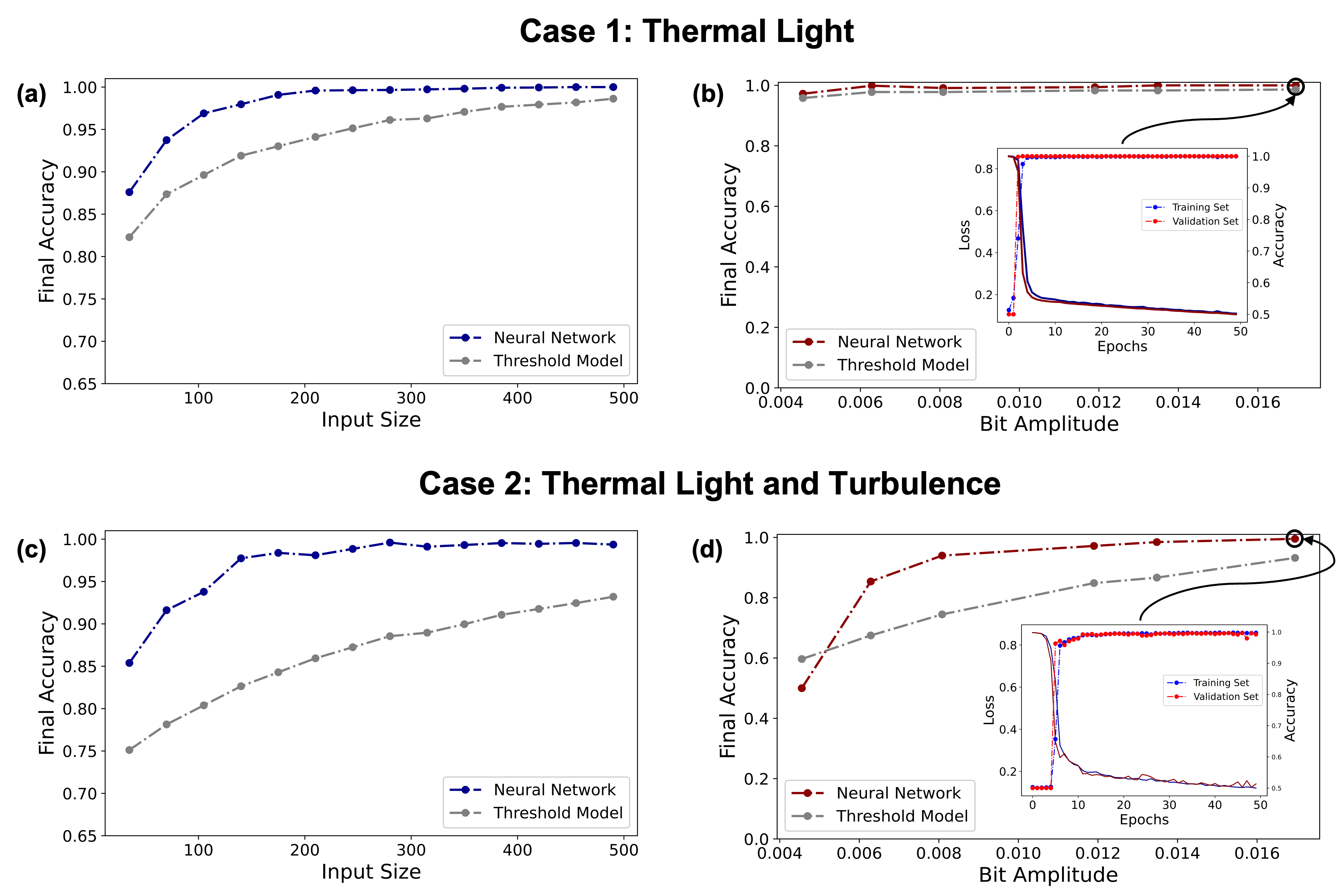}
         \caption{Results from the first neural network. (a) and (b) Final classification accuracy for varying input sizes for Case 1: Thermal light and Case 2: Thermal light and turbulence, respectively. The input size corresponds to the amount of consecutive values input into the CNN, where 35 consecutive values is equivalent to 1 bit. The accuracy of the neural network is compared to a threshold model. We found that higher input sizes lead to better performance in detecting the coherent bit sequences. The input size 490 achieved the highest accuracy and was used to model the final classification accuracy for a range of SNRs in Figure (c) and (d). This was similarly varied for both noise cases and compared to a threshold method. (inset) The accuracy and loss while running the neural network for the highest SNR. The red lines are the validation set and the blue lines are the training set. The network was run for 50 epochs for each data point. }
         \label{FinalNN1}
\end{figure*}

Another advantage to deep learning is the capability to change input topology of the data to improve classification. In order to optimize the accuracy of our initial neural network, we vary the size of the individual matrices input into the neural network. As one coherent OOK bit pulse is defined as an increment of 35 values, we began by inputting one bit at a time, and the neural network was run as outlined in Figure \ref{NN1} for 50 epochs to determine if there was a bit present. We increase the size of the input matrix by 35 consecutive values at a time until we reached 490 consecutive values, or 14 bit sequences.  The resulting final accuracy for different input topologies for the first neural network is shown in Figure \ref{FinalNN1} (a) and (c). Additionally, the neural network was tested for varying bit amplitudes. The bit amplitude relates directly to the power of the signal, which we varied relative to a steady noise for the two cases of thermal noise and thermal noise and turbulence. As the bit amplitude is varied relative to a common noise, the bit amplitude is proportional to the SNR. The SNR was varied using 14 bit sequences at a time, as this resulted in the highest classification accuracy. Our results are shown in Figure \ref{FinalNN1} (b) and (d).

The network achieved higher accuracy as we increased the number of bit values input into the neural network at a time for coherent OOK bits in both noise cases, and the input sizes of 385 and higher had a classification accuracy above 98\% for both the training and validation sets. The accuracy of our neural network is compared to a threshold model which, classifies an input sequence by comparing its mean amplitude to the mean amplitude value of all data, establishing a threshold for which the data is compared. In this case, if a given increment of data has a higher average intensity than the average intensity of an entire corresponding data set, the threshold method predicts that coherent bits are present. Similar threshold models have been implemented, including more complex adaptive threshold methods \cite{Thr1, Thr2}. The neural network outperformed the threshold model in both scenarios. 
The random generation of coherent OOK bit streams, in addition to thermal background, can lead to long sequences of bits with no ``ON'' bits present, or a long stream of ``OFF'' bits.  This mimics the data set containing thermal background only. This is reflected by the low accuracy in classification for smaller input sizes. Longer input sizes for the training set lead to a low probability of a batch with no `ON' bit pulses. In addition, neural networks need large amounts of data to train a model. Given these results, long input sequences are recommended for training the neural networks. We find the highest accuracy  for a range of SNRs occurs when using 490 bits as the input size and proceed with this network architecture. 

\begin{figure*}[htb]
\centering
\includegraphics[width=.8\linewidth]{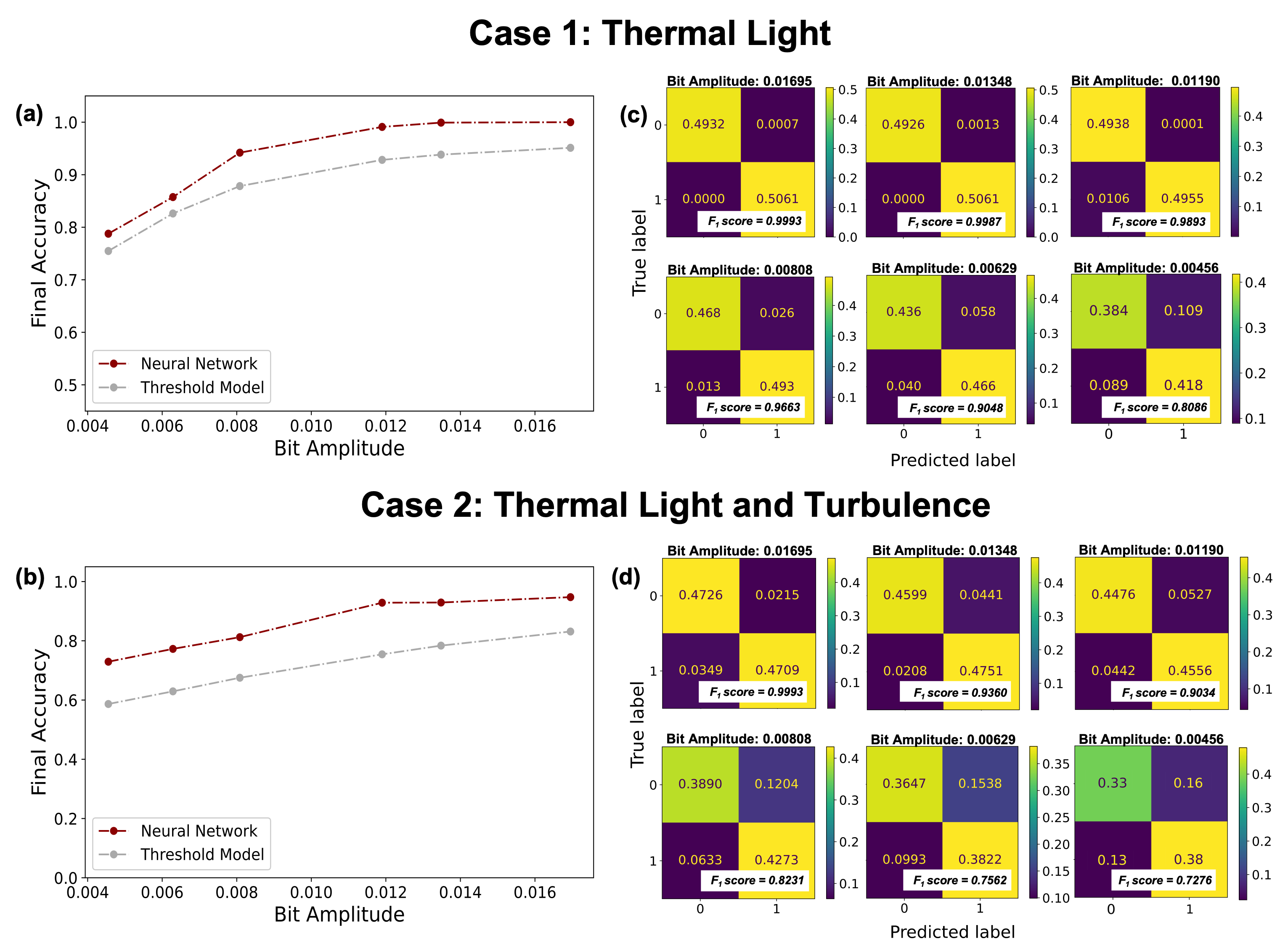}
    \caption{Results from the second neural network. (a) and (b) Final classification accuracy for varying SNR. The neural network classifies and demodulates the coherent bit signal. The final results is again compared to a threshold model which is in gray. (b) and (d) Confusion matrices for the normalized true versus predicted output of the CNN. This was run for each SNR. The F$_1$ value is a statistical tool for binary classification, which is a measure of a test's accuracy. Thermal light achieved notably higher accuracy and F$_1$ scores than thermal light and turbulence combined, however the CNN performed well in both cases for higher SNR. In addition, as the SNR decreased, we find a higher likelihood for the neural network to improperly classify ``OFF'' as ``ON''.}
\label{FinalNN2}
\end{figure*}

Upon determining the presence of coherent bits, the data is input into the second neural network. The results from the second neural network are shown in Figure \ref{FinalNN2}. The results for vaying SNR are compared to the fixed threshold method. Here, the threshold method works by taking the average intensity of an entire data set that contains coherent bits. If the intensity of a given bit is greater than the average intensity, the threshold method predicts the bits is a ``ON'', and if the bits has a lower intensity than the average intensity, the bit is predicted to be ``OFF''. In both scenarios of thermal noise only and thermal noise and turbulence, the classification accuracy using the neural networks outperformed the threshold method. 

Confusion matrices were also generated at each SNR after running the CNN to determine the amount of improperly identified bits were present. The F$_1$ value was also calculated which is given by:

\begin{equation}
    \text{F}_1 = \frac{\text{TP}}{\text{TP}+\frac{1}{2}(\text{FP}+\text{FN})}
\end{equation}

where TP corresponds to ``OFF'' bits being correctly classified, FP corresponds to ``ON'' bits being classifed as ``OFF'', and FN corresponds to ``OFF'' being classified as ``ON'', and the closer to 1 the F$_1$ score, the better the binary classification. As we run the CNN for lower SNR, the CNN is more likely to missclassify ``OFF'' bits as ``ON'', due to the high noise that is nearing the amplitude of the coherent bit. 

The resulting BER is defined as BER = $1-$Final Accuracy, where the BER using our convolutional neural network is near-zero for high SNRs. These results demonstrate the advantage of using neural networks in communication schemes for power limited or noisy scenarios, where we have nearly 100\% accuracy for high SNRs. In addition, adaptive optics has capabilities to effectively raise the signal-to-noise ratio, including recent advancements showing adaptive optics capability to increase the SNR for a range of channel radiances and turbulences for quantum FSO communication \cite{FSO, AO1}. Our results using the neural network are presented with no other demodulation technique implemented in conjunction with our neural network. If used with adaptive optics, this would allow higher accuracy for noisy environments, as well as the capability to extend along the horizon for future free space optical communications. Utilizing a pre-trained machine learning scheme will not effect cost, size, weight or power for future use in free space optical communication, and is recommended to be used with other demodulation schemes. This may also expand the scope of the demodulation to other non-uniform loss affecting the bits prior to arrival at the receiver, as statistics of noisy environments need not be known to neural networks to classify bit sequences. 

\section{Discussion}
Reliable free-space optical communication systems can operate at high power levels, transmit large data rates, and allow for long distance communication. One of the most significant obstacles to FSO communication schemes is that the effects of noisy environments due to atmospheric transmission, turbulence,  weather conditions, and thermal background noise affect the signal and resulting demodulation. As we aim to communicate over longer distances, in noisier environments, to support high data transmission and data reliability, the necessity for advanced optical communication is clear. We demonstrate the usefulness of CNNs in decreasing the bit error rate in post-processing and drastically increasing the classification accuracy over threshold demodulation. The use of demodulation techniques utilizing neural networks allows us to significantly decrease the overhead and cost needed to demodulate in extremely noisy environments. One benefit of this approach is that it may be combined with current error correction and demodulation techniques, as it is implemented as a post-processing technique. The use of neural networks with adaptive optics allow us to expand optical communication further down the horizon. Further work will allow us to look at other modulation schemes such as pulse-position modulation and quadrature phase-shift keying modulation, and optimize the computational time needed to train the neural networks. 

Additionally, the system may be pre-trained prior to deployment, allowing for rapid, near real-time implementation. Meeting the increasing need for optical devices for communication is a challenge, and deep learning models allow us to expand the range of uses for optical communication and increase information capacity in such systems. We hope this work can be used to enhance satellite communication in the near term and expand to demonstrations in quantum optical communications \cite{Q1,Q2}.

\section*{Funding} U. S. Office of Naval Research (N000141912374); National Science Foundation Graduate Research Fellowship (2139911).

\section*{Acknowledgments} M.B. would like to acknowledge NASA Goddard and the NSF Graduate Research Fellowship program. P.R. and H.L. would like to acknowledge the US Air Force Office of Scientific Research and the US Army Research Office.

\section*{Data Availability Statement}

The data that support the findings of this study are available from the corresponding author, M.B., upon reasonable request.

\bibliography{sample}

\end{document}